\documentclass[runningheads]{llncs}
\usepackage[T1]{fontenc}
\usepackage{graphicx}
\usepackage{color}
\usepackage{url}            
\usepackage{booktabs}       
\usepackage{array}
\usepackage{multirow}
\usepackage{amsfonts}       
\usepackage{amsmath}        %
\usepackage{amssymb}        %
\usepackage{subcaption}
\usepackage{lipsum}

\begin{document}
\title{Automatic Cephalometric Landmark Localization on CBCT-Derived Digitally Reconstructed Radiographs for Skeletal Malocclusion Classification}
\titlerunning{Landmark Localization for Skeletal Malocclusion Classification}
%
\author{
Benjamin Hou\inst{1} \and
Konstantinia Almpani\inst{2} \and
Janice S. Lee\inst{2} \and 
Zhiyong Lu\inst{1}
}
\authorrunning{B. Hou et al.}
%
\institute{
Division of Intramural Research, National Library of Medicine \and
Craniofacial Anomalies and Regeneration Section, \\
National Institute of Dental and Craniofacial Research, \\
National Institutes of Health, Bethesda, MD, USA.
}
\maketitle              
\begin{abstract}
Manual cephalometric landmark annotation is important for craniofacial assessment but is labor-intensive and difficult to scale. We introduce CephViT, a Vision Transformer-based model for automated 2D lateral cephalometric landmark localization, and evaluate its use in downstream skeletal malocclusion classification. CephViT was trained and benchmarked on a public lateral cephalogram dataset, achieving a mean radial error of 1.28 $\pm$ 1.42 mm and a successful detection rate of 92.0\% at 3.0 mm. Because the private evaluation cohort consisted of 3D CBCT scans, lateral cephalogram-like digitally reconstructed radiographs (DRRs) were generated from each volume and used as 2D inputs to the landmark localization model. Landmark coordinates were normalized into a common coordinate frame, and skeletal malocclusion classification was performed using landmarks shared between the reference and DRR-based pipelines. Classification performance using DRR-localized landmarks was comparable to that obtained using manually annotated reference landmarks, with accuracies of 70.0\% and 68.3\%, respectively. These results support the feasibility of automated cephalometric analysis on CBCT-derived DRRs for skeletal malocclusion assessment.
\keywords{Cephalometric landmark detection \and DRR \and Cone-beam CT \and Skeletal malocclusion \and Vision Transformer}
\end{abstract}
\section{Introduction}

Skeletal malocclusion is a clinically important manifestation of aberrant craniofacial development, which has been associated with a broad range of comorbidities spanning orofacial, respiratory, musculoskeletal, neuropsychological, and systemic domains \cite{koskela2021relation,nowak2025associations,angelo2024association,nicot2020condyle,togawa2009gastroesophageal,alrashed2021relationship}. Early and accurate diagnosis can support growth monitoring, guide interceptive orthodontic treatment management, and help distinguish cases that may later require more complex orthopedic or surgical treatment \cite{zhou2024expert,zhou2025expert,lin2025expert}. Historically, standard cephalometric analysis on lateral cephalograms has long played a central role in this process by quantifying an individual’s potential degree of deviation from the normative state, as defined by cephalometric measurement values collected from individuals with no skeletal malocclusion (orthognathic). These are linear and angular measurements that are used to assess the spatial relationships of the cranial base, maxilla, mandible, and dentition, thereby helping define the anatomical basis of malocclusion, and assisting in tracking craniofacial growth and development over time \cite{mcnamara1984method,bjork1983normal}. 

In modern clinical practice, Cone-Beam Computed Tomography (CBCT) is increasingly used in the craniofacial field, offering a complete and more accurate three-dimensional (3D) assessment of craniofacial anatomy, particularly in cases involving facial asymmetry, complex skeletal discrepancies, or treatment planning beyond what can be reliably inferred from a two-dimensional (2D) cephalogram. In addition, inherent problems of standard lateral cephalograms, including head positioning errors, magnification and distortion, and superimposition of bilateral structures can significantly undermine the accuracy of landmark annotation on these images. 

CBCT-based cephalometric analysis still relies heavily on expert landmark annotation despite reduced superimposition and more accurate visualization of anatomy \cite{almpani2023three}. Most automated cephalometric landmark detection models, however, have been developed for 2D lateral cephalograms \cite{jiang2023artificial}. Recent work has shown that deep learning is a promising approach for automated cephalometric landmark detection, but robustness, generalizability, and sufficient coverage of clinically relevant landmark sets remain important challenges \cite{schwendicke2021deep,hendrickx2024can,gillot2023automatic}.

Digitally reconstructed radiographs (DRRs) provide a radiograph-like 2D representation of CBCT volumes and can therefore enable established 2D cephalometric landmark detectors to be applied to CBCT-derived data. Therefore, the aim of this study was to evaluate whether skeletal malocclusion classification can be assessed from automatically localized cephalometric landmarks on CBCT-derived DRRs. A Vision Transformer-based model, CephViT, was trained on a public lateral cephalogram dataset and applied to DRRs generated from a private CBCT cohort, with downstream skeletal malocclusion classification compared against a reference pipeline based on manual CBCT landmark annotations.

\section{Materials and Methods}

\begin{figure}[!ht]
    \centering
    \includegraphics[height=2.6cm]{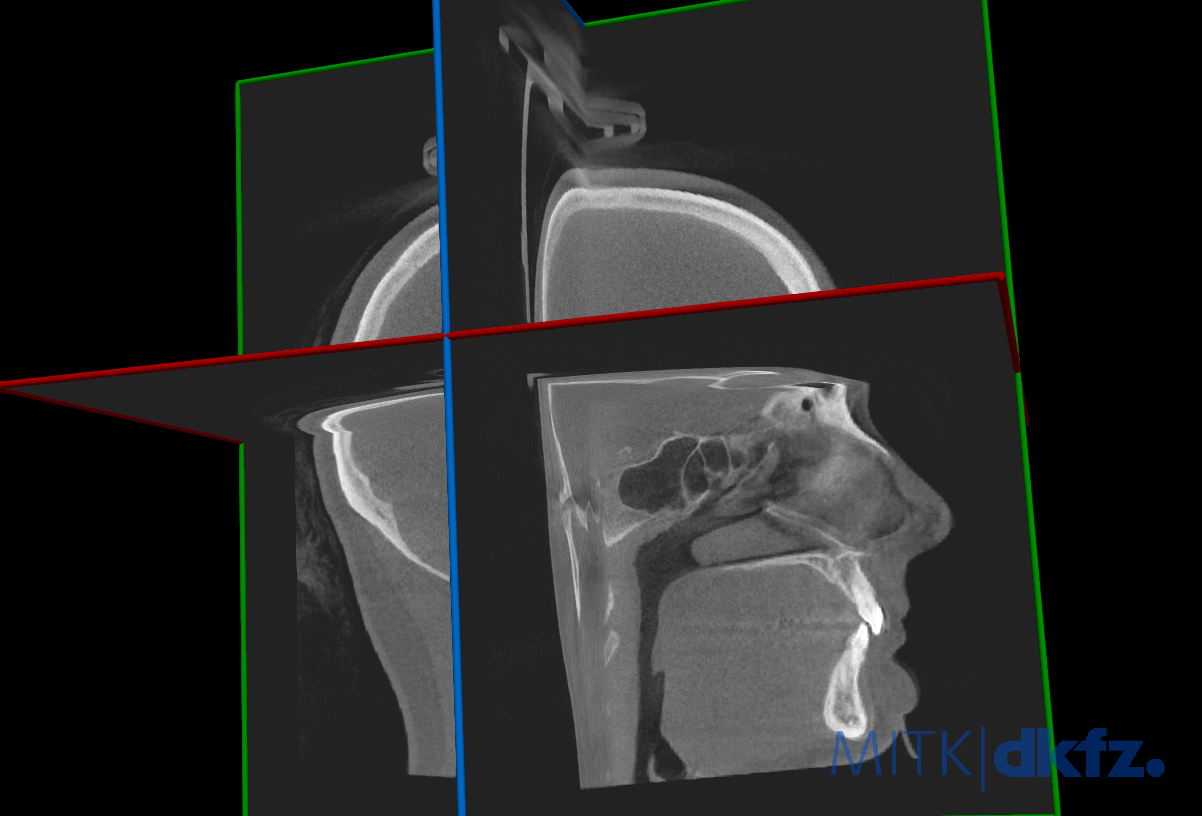}
    \includegraphics[height=2.6cm]{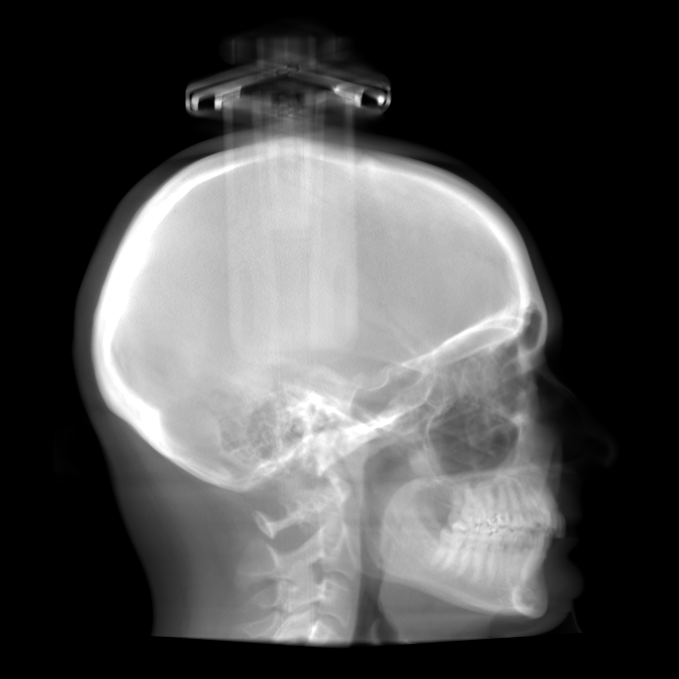}
    \includegraphics[height=2.6cm]{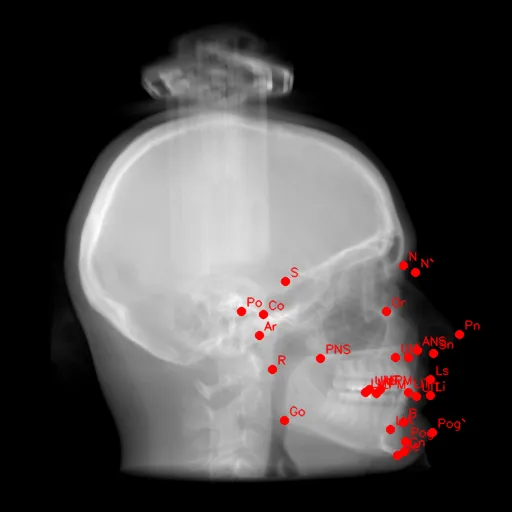}
    \includegraphics[height=2.6cm]{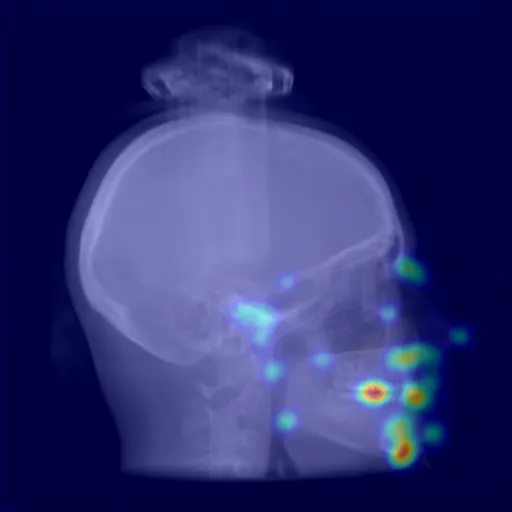}
    \caption{Overview of the study pipeline. From L-to-R: CBCT volume, lateral DRR, cephalometric landmark localization using CephViT, landmark heatmap output. The localized landmark coordinates were subsequently normalized and used for downstream skeletal malocclusion classification.}
    \label{fig:pipeline}
\end{figure}

This study comprised two linked components: automatic cephalometric landmark localization and downstream skeletal malocclusion classification. CephViT, a landmark localization network, was developed using the Aariz lateral cephalogram dataset \cite{khalid2025benchmark,khalid2022cepha29} and subsequently applied to DRRs generated from a separate CBCT dataset (Fig. \ref{fig:pipeline}). The CBCT dataset included both manual reference landmark annotations and skeletal malocclusion labels, enabling comparison of malocclusion classification using automatically localized versus reference landmarks. Model and accompanying code are available at: 
\\ \url{https://huggingface.co/nlm-dir/CephViT},
\\ \url{https://github.com/farrell236/midas-journal-784}.

\subsection{Datasets}

This retrospective study involved both public and private datasets. The private dataset was deidentified prior to analysis, and its use was approved by the institutional review board with a waiver of informed consent. 

The public dataset, Aariz, consisted of 1,000 lateral cephalometric radiographs from patients aged 12--62 years \cite{khalid2025benchmark,khalid2022cepha29}. Images were acquired using seven imaging devices with varying resolutions and annotated with 29 cephalometric landmarks, comprising 15 skeletal, 8 dental, and 6 soft-tissue landmarks. Annotations were made in a two-phase process involving junior and senior orthodontists, with the final ground truth defined as the average of both annotations. 

The private CBCT dataset, 
NIDCR-CARS (IRB \#16-D-0040), 
consisted of 300 scans from 300 patients. Age and sex were available for 297 patients (134 male, 163 female), who were aged 13.6--65.4 years (mean age, 24.2 years \(\pm\) 12.8). Skeletal landmarks were annotated on the CBCT volumes using a validated 3D annotation protocol \cite{almpani2023three}, with 43 landmarks spanning the face, maxilla, mandible, and cranial base. Each subject was assigned a skeletal malocclusion label of Class I, Class II, Class IIIA, or Class IIIB, yielding an analysis-cohort distribution of 99, 99, 27, and 75 subjects, respectively.

\subsection{DRR generation from CBCT}

DRRs were generated from the CBCT volumes using the Siddon-Jacobs ray-tracing algorithm \cite{siddon1985fast}, which produces radiograph-like projection images by tracing rays from a virtual x-ray source to a detector plane and integrating voxel values along each ray path through the 3D volume. Formally, the DRR intensity at detector location \((u,v)\) can be written as:
\begin{equation}
    D(u,v) = \int_{\mathcal{L}(u,v)} \mu(\mathbf{r}) \, d\mathbf{r},
\end{equation}
where \(D(u,v)\) denotes the projected image intensity, \(\mathcal{L}(u,v)\) is the ray path from the virtual x-ray source to detector coordinate \((u,v)\), and \(\mu(\mathbf{r})\) is the voxel intensity at spatial location \(\mathbf{r}\) in the CBCT volume. DRR generation was performed using the ITK-based Siddon-Jacobs ray-casting implementation described by Wu et al. \cite{wu2010itk}. Projection images were generated at a resolution of \(512 \times 512\) pixels with detector pixel spacing of \(0.51 \times 0.51\) mm and a source-to-isocenter distance of 1000 mm. To suppress contributions from air and low-density background regions, voxels with intensities below -900 were excluded during projection.

\subsection{Automatic cephalometric landmark localization}

Automatic cephalometric landmark localization was performed using a Vision Transformer (ViT)-based heatmap regression model \cite{dosovitskiy2020image}. Given an input image \(I \in \mathbb{R}^{H \times W \times 3}\), the network \(f_\theta\) produced a set of output landmark heatmaps, denoted by \(\hat{H}\):
\begin{equation}
    \hat{H} = f_\theta(I), \qquad \hat{H} \in \mathbb{R}^{N \times h \times w},
\end{equation}
where \(N\) denotes the number of landmarks and \(h \times w\) is the heatmap resolution. The model used a ViT backbone with \(16 \times 16\) patches and \(512 \times 512\) pixel inputs. The classification token was discarded, and the remaining patch embeddings were reshaped into a 2D feature map. A \(1 \times 1\) convolution projected the transformer features to 256 channels, followed by a convolutional decoder with two bilinear upsampling stages and intermediate \(3 \times 3\) convolution layers with ReLU activation. A final \(1 \times 1\) convolution generated one heatmap per cephalometric landmark. Supervision was formulated as heatmap regression. For each landmark \(n\), a Gaussian target heatmap \(H_n\) was generated as:
\begin{equation}
    H_n(x,y)=\exp\!\left(-\frac{(x-x_n)^2+(y-y_n)^2}{2\sigma^2}\right),
\end{equation}
where \((x_n,y_n)\) is the landmark location in heatmap space and \(\sigma\) controls the Gaussian spread. The full set of target heatmaps is denoted by \(H\). The model was trained by minimizing the mean squared error between the output heatmaps \(\hat{H}\) and the target heatmaps \(H\):
\begin{equation}
    \mathcal{L}_{\mathrm{MSE}}=\mathrm{MSE}(\hat{H},H).
\end{equation}
At inference, landmark coordinates were recovered from the output heatmaps using DARK-style decoding with local refinement \cite{zhang2020distribution} and then mapped back to image space for evaluation.

The model was trained for 100 epochs with a batch size of 16 using AdamW, with an initial learning rate of $1\times10^{-4}$ and weight decay of $1\times10^{-4}$. Gaussian target heatmaps were generated with $\sigma=2.0$. On-the-fly augmentation included affine transforms (scale, $0.95-1.05$; translation, $\pm3$\%; rotation, $\pm8^\circ$; shear, $\pm3^\circ$), random brightness and contrast adjustment, and Gaussian noise. Training was performed on an HPC cluster node equipped with an AMD EPYC 7543P CPU and an NVIDIA A100 GPU, full training completed in approximately 5 hours.

\subsection{Landmark-based skeletal malocclusion classification}

As the reference landmarks were annotated in 3D CBCT space rather than on 2D radiographs, each subject's 3D landmark configuration was transformed into a common anatomical frame and projected onto a lateral sagittal plane for downstream analysis.

To achieve this, a cranial-base centroid defined the origin, while bilateral craniofacial landmarks and the posterior-to-anterior palatal axis defined consistent left-right, anterior-posterior, and superior-inferior axes. Bilateral landmarks were collapsed to their midpoints where appropriate, and a final Nasion-Sella scale normalization reduced residual inter-subject size variation.

Cephalometric landmarks automatically localized on CBCT-derived DRRs were normalized in 2D by aligning each image-plane landmark configuration to a representative reference shape using a similarity transformation estimated from the 12 common cephalometric landmarks. The transform was then applied to the full localized landmark set before feature extraction.

For direct comparison between the reference and DRR-localized landmark pipelines, skeletal malocclusion classification was performed using coordinate and pairwise-distance features derived from the 12 cephalometric landmarks common to both representations: A, ANS, Ar, B, Go, Me, N, Or, PNS, Po, Pog, and Sella. 

\section{Results}

\subsection{CephViT Landmark Localization Performance}

On the held-out Aariz test set, CephViT achieved a mean radial error of 1.28 mm \(\pm\) 1.42 mm across all 29 landmarks. Successful detection rates were 82.1\%, 87.9\%, 92.0\%, and 95.7\% at 2.0, 2.5, 3.0, and 4.0 mm, respectively. Landmark-wise results are summarized in Table \ref{tab:aariz_comparison}. Compared with previously reported methods evaluated on the same benchmark split, CephViT achieved competitive localization performance, with lower mean radial error and a smaller error spread, together with higher or comparable successful detection rates across the evaluated thresholds. 

\begin{table}[!t]

\centering
\caption{Cephalometric landmark detection performance on the Aariz test set.}
\label{tab:aariz_comparison}
\scalebox{0.95}{
\begin{tabular}{@{}>{\raggedright\arraybackslash}p{3cm}
                   >{\centering\arraybackslash}p{2.2cm}
              *{4}{>{\centering\arraybackslash}p{1.4cm}}@{}}
\toprule
~~\multirow{2}{*}{Method} &
\multirow{2}{*}[-0.3em]{\shortstack[c]{MRE $\pm$ SD \\ (mm)}} &
\multicolumn{4}{c}{SDR (\%)} \\ 
\cmidrule(l){3-6}
& & 2.0 mm & 2.5 mm & 3.0 mm & 4.0 mm \\
\midrule
~~Khan et al. \cite{khan2024enhancing}   & 1.92 $\pm$ 7.85 & 78.54 & 85.72 & 89.64 & 94.49 \\
~~Khalid et al. \cite{khalid2024two}     & 1.87 $\pm$ 4.01 & 75.17 & 82.43 & 88.78 & 93.01 \\
~~Khan et al. \cite{khalid2025benchmark} & 1.69 $\pm$ 3.36 & 81.18 & 87.28 & 90.82 & 94.82 \\
\midrule
~~\textbf{CephViT (ours)} & \textbf{1.28 $\pm$ 1.42} & \textbf{82.10} & \textbf{87.90} & \textbf{92.00} & \textbf{95.70} \\
\bottomrule
\end{tabular}
}
\vspace{-2mm}
\end{table}

\subsection{Malocclusion Classification from Reference Landmarks}

Reference landmark normalization and projection were successful for all 300 subjects, with no exclusions at this stage. The projected landmark configurations formed a coherent anatomical space; Sella, Porion, Articulare, Orbitale, and posterior nasal spine showed the lowest mean radial deviation from the cohort mean shape, whereas Menton, Pogonion, B point, Gonion, and A point showed the greatest variability. A post hoc Tukey-rule screen identified 11 potential shape outliers, although no explicit outlier exclusion was applied in the reference-landmark analysis.

\begin{figure}[!b]
    \centering
    \includegraphics[height=4.8cm]{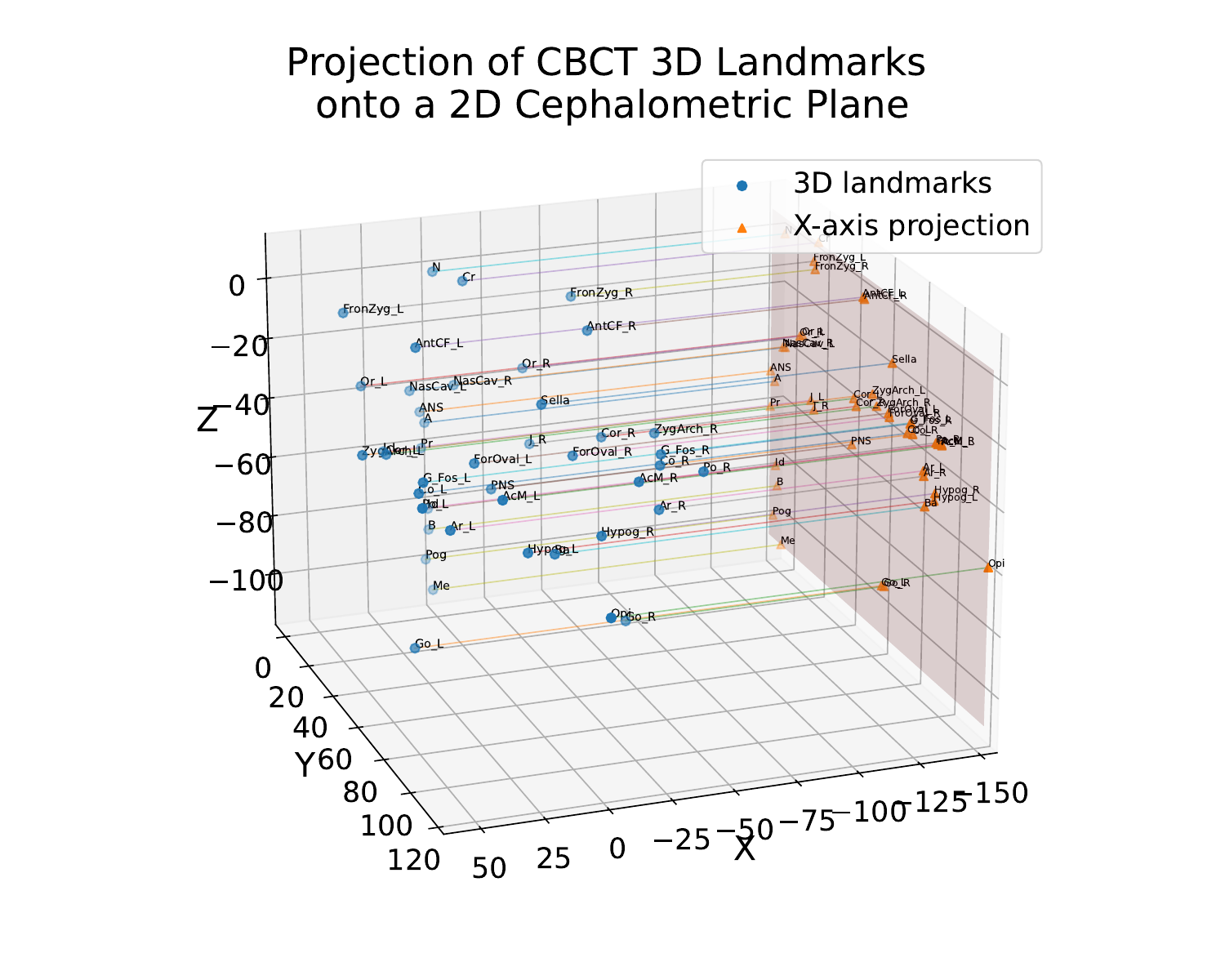}
    \includegraphics[height=4.8cm]{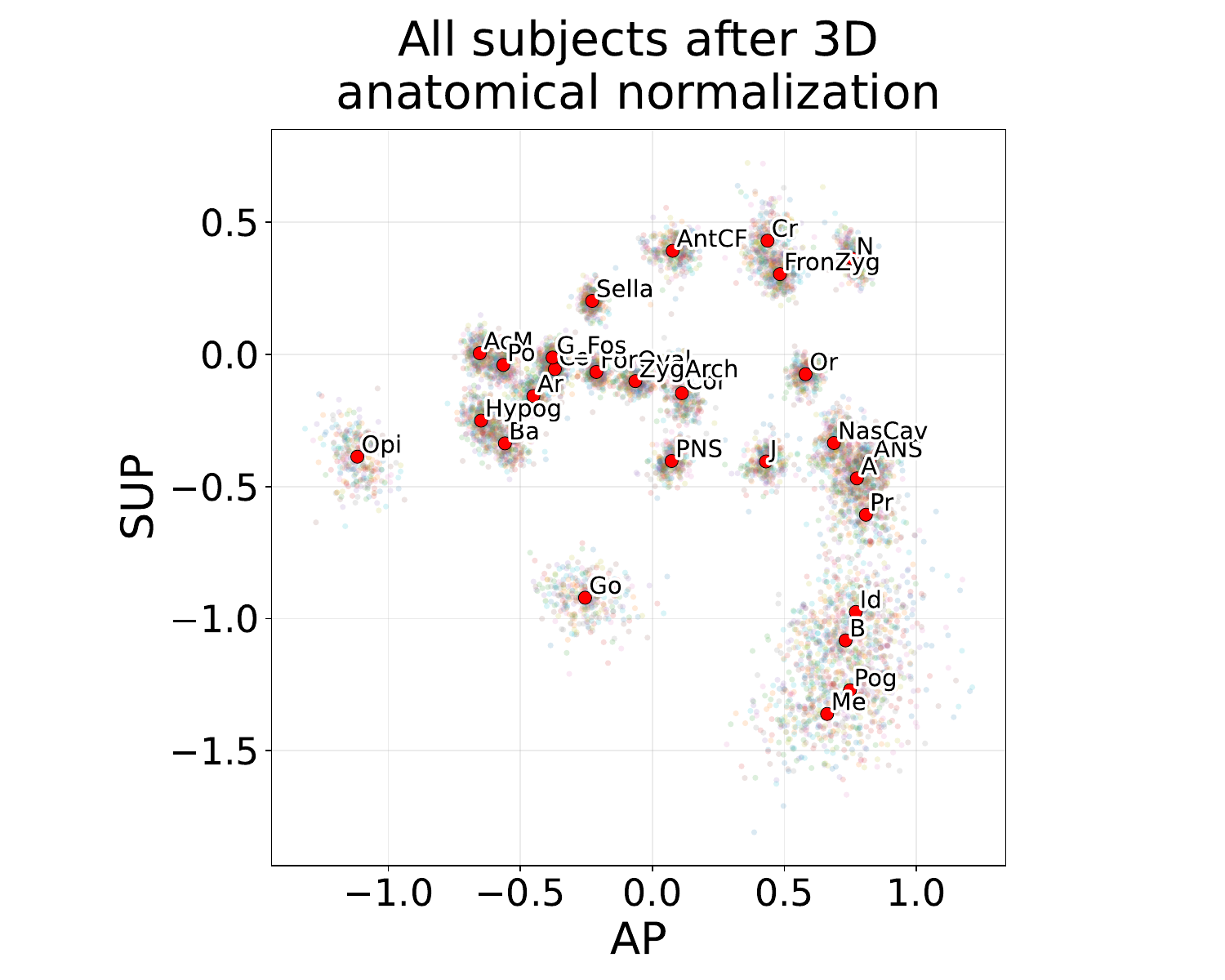}
    \caption{Projection and normalization of CBCT-derived reference landmarks. Left; 3D CBCT landmark configuration and lateral projection onto a 2D cephalometric plane. Right; all projected subjects after 3D anatomical normalization, with mean landmark positions shown in red.}
    \label{fig:ref_landmark}
\end{figure}

Skeletal malocclusion classification was then performed using the normalized two-dimensional landmark coordinates. Performance was evaluated using a multinomial logistic regression model with mean imputation, feature standardization, balanced class weights, and 5-fold stratified cross-validation, with class-wise metrics computed from out-of-fold predictions.

Using the normalized reference landmark coordinates, skeletal malocclusion classification achieved an overall accuracy of 68.3\%. Class-wise F1 scores were 0.667 for Class I, 0.768 for Class II, 0.431 for Class IIIA, and 0.703 for Class IIIB. Performance was strongest for Class II and Class IIIB, whereas Class IIIA showed the lowest precision and F1 score, consistent with greater difficulty in separating this smaller subgroup. The macro-averaged F1 score was 0.642, and the weighted-average F1 score was 0.688 (Table \ref{tab:ref_landmark_classification}).

\subsection{Malocclusion Classification from DRR-Localized Landmarks}

\begin{figure}[!b]
    \centering
    \includegraphics[height=4.6cm]{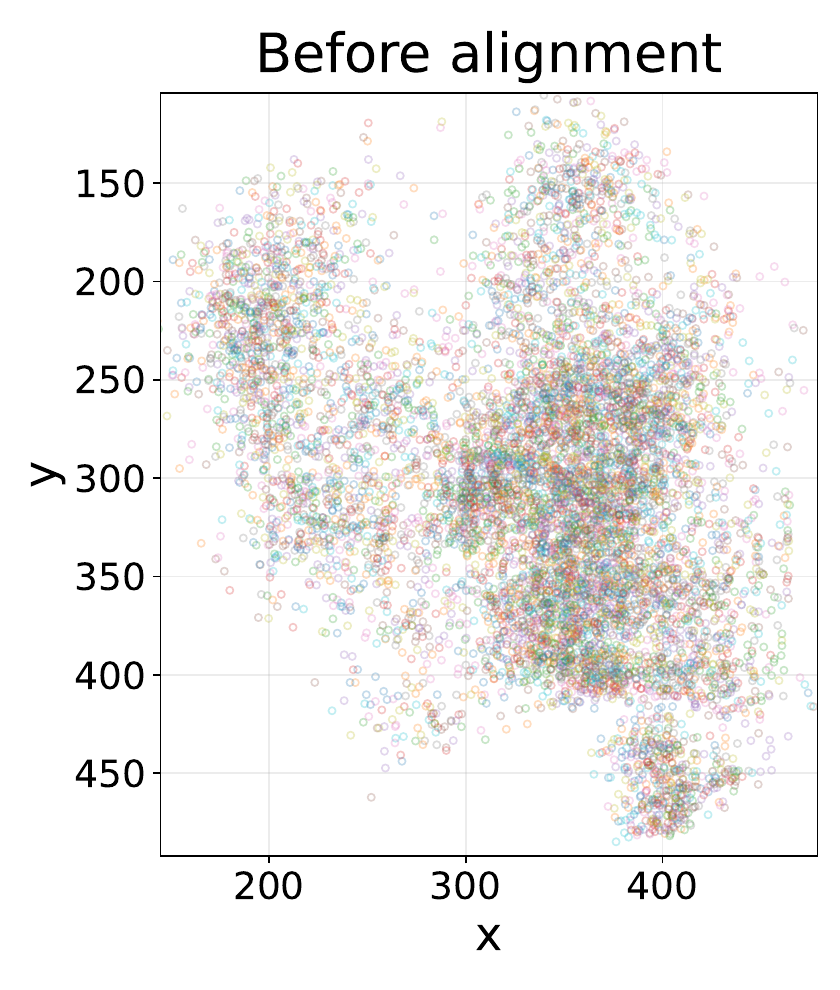} \hspace{2em}
    \includegraphics[height=4.6cm]{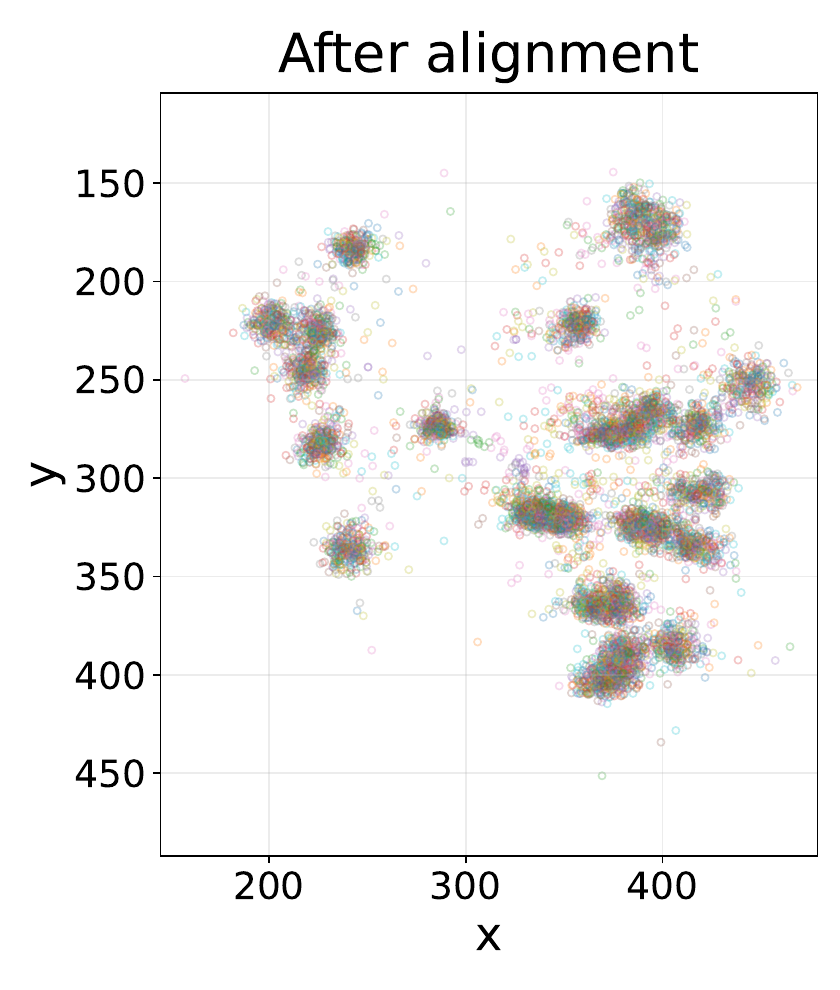}
    \caption{2D normalization of DRR-localized cephalometric landmarks before (left) and after (right) similarity-based alignment to a common coordinate system.}
    \label{fig:drr_landmark}
\end{figure}

For the DRR-based landmark pipeline, all 300 manifest-common subjects were retained. Automatically localized 2D landmark configurations were aligned to a representative reference configuration using a similarity transformation estimated from the 12 common cephalometric landmarks, with no post-registration case exclusion (Figure \ref{fig:drr_landmark}).

After alignment, the most stable landmarks were posterior nasal spine, A point, Orbitale, Articulare, and Gonion, whereas the most variable were Sella, Nasion, Menton, Porion, and Pogonion. A post hoc Tukey-rule screen of subject-level mean landmark deviation identified 36 potential shape outliers in the aligned DRR cohort. Classification used the same stratified 5-fold cross-validation fold assignments as the 3D reference-landmark pipeline to enable a fair paired comparison.

Using the normalized DRR-localized landmark coordinates, skeletal malocclusion classification achieved 70.0\% accuracy, macro-F1 of 0.647, and weighted-F1 of 0.708 under 5-fold cross-validation (Table \ref{tab:drr_landmark_classification}). Paired significance testing was performed using an exact McNemar test, while macro- and weighted-F1 differences were assessed using paired permutation testing with bootstrap 95\% confidence intervals. No statistically significant differences were observed for accuracy (68.3\% vs. 70.0\%; \(p=0.696\)), macro-F1 (0.642 vs. 0.647; \(p=0.894\)), or weighted-F1 (0.688 vs. 0.708; \(p=0.555\)), supporting comparable performance of the DRR-based method relative to the reference-landmark standard. The small numerical advantage of the DRR-localized pipeline should not be interpreted as superiority and may reflect cross-validation variability or effects of the shared 2D projection.

\begin{table}[!ht]
\vspace{-1em}
\centering
\caption{Five-fold cross-validated malocclusion classification.}
\label{tab:ref_landmark_classification}
\label{tab:drr_landmark_classification}
\scalebox{0.95}{%
\begin{tabular}{@{}lcccccc@{}}
\toprule
      & Support & \begin{tabular}[c]{@{}c@{}}Reference \\ landmarks\end{tabular} & \begin{tabular}[c]{@{}c@{}}DRR-localized \\ landmarks\end{tabular} & $\Delta$ & \begin{tabular}[c]{@{}c@{}}95\% CI \\ for $\Delta$\end{tabular} & Paired $p$ \\ \midrule
Accuracy    & 300     & 0.683                                                          & 0.700                                                              & +0.017   & {[}-0.050, 0.083{]}                                             & 0.696      \\
Macro F1    & 300     & 0.642                                                          & 0.647                                                              & +0.005   & {[}-0.066, 0.074{]}                                             & 0.894      \\
Weighted F1 & 300     & 0.688                                                          & 0.708                                                              & +0.020   & {[}-0.045, 0.084{]}                                             & 0.555      \\ \midrule
Class I     & 99      & 0.667                                                          & 0.691                                                              & +0.024   & {[}-0.071, 0.119{]}                                             & --         \\
Class II    & 99      & 0.768                                                          & 0.794                                                              & +0.026   & {[}-0.052, 0.107{]}                                             & --         \\
Class IIIA  & 27      & 0.431                                                          & 0.364                                                              & -0.067   & {[}-0.223, 0.080{]}                                             & --         \\
Class IIIB  & 75      & 0.703                                                          & 0.740                                                              & +0.036   & {[}-0.065, 0.139{]}                                             & --         \\ \bottomrule
\end{tabular}%
}
\vspace{1mm}
\begin{flushleft}
\footnotesize
Values are computed from out-of-fold predictions. $\Delta$ denotes DRR-localized minus reference-landmark performance. Confidence intervals were estimated using paired bootstrap resampling (\(n=10{,}000\)). Paired $p$ values were computed using an exact McNemar test for accuracy and paired permutation tests for macro- and weighted-F1 (\(n=100{,}000\)). Class-wise F1 rows are not separately hypothesis-tested.
\end{flushleft}
\vspace{-1em}
\end{table}

\section{Discussion}

This study evaluated whether CBCT-derived DRRs could support automated 2D cephalometric landmark localization and downstream skeletal malocclusion classification. This workflow is intended as a practical bridge, not a replacement for native 3D CBCT landmarking. When CBCT volumes are available but no validated 3D model exists for the full 43-landmark reference protocol, DRRs allow established 2D cephalometric landmark localizers to be applied to CBCT-derived data for downstream classification. Using the same 300 subjects and identical cross-validation folds, the DRR-localized landmark pipeline achieved performance comparable to the reference pipeline based on manual 3D CBCT landmarks projected to 2D (accuracy, 70.0\% vs. 68.3\%; macro F1, 0.647 vs. 0.642). These findings suggest that DRR-localized cephalometric landmarks preserve sufficient craniofacial geometry for skeletal subclass discrimination.

Automated extraction of cephalometric geometry from CBCT-derived DRRs could support scalable craniofacial phenotyping or pre-screening, particularly in settings where manual CBCT landmark annotation is impractical. Because manual 2D annotations were not available for the generated DRRs, domain transfer was assessed indirectly through downstream classification rather than direct DRR landmark error. Nevertheless, the DRR-based workflow remains dependent on image quality, landmark localization accuracy, and geometric normalization. Greater landmark dispersion and subject-level shape outliers after alignment indicate that robust localization and registration remain important failure modes to address.

Future work should validate the pipeline in larger external cohorts across CBCT acquisition protocols, fields of view, image quality, and DRR-generation settings, and assess whether image-based or hybrid image-landmark models can further improve differential classification between skeletal Class III subgroups.

\begin{credits}
\subsubsection{\ackname} This research was supported by the Intramural Research Program of the National Institutes of Health (NIH). The contributions of the NIH authors are considered Works of the United States Government. The findings and conclusions presented in this paper are those of the authors and do not necessarily reflect the views of the NIH or the U.S. Department of Health and Human Services. This work utilized the computational resources of the NIH High-Performance Computing Biowulf cluster (\url{https://hpc.nih.gov}).

\end{credits}

\bibliographystyle{splncs04}
\bibliography{references}

\newpage

\section*{Supplementary Material}

\subsection*{Malocclusion Classification}

The most common current skeletal classification is based on the position of the maxilla and mandible relative to the anterior cranial base in the sagittal plane [A]. This classification is based on the results of the cephalometric analysis. When both the maxilla and mandible are in positions that are within the normal range of the existing cephalometric measurement normative values, close to the position of the anterior cranial base in the sagittal plane, there is no skeletal malocclusion, and the skeletal classification is Class I (orthognathic). When the maxilla is in a more anterior position and/or the mandible is in a more posterior position in relation to the anterior cranial base, the resulting skeletal malocclusion is classified as Class II, commonly described as ``retrognathic'' or ``overbite''. When opposite relationships exist, with the maxilla in a more posterior position and/or the mandible in a more forward position in relation to the anterior cranial base the resulting malocclusion is classified as Class III, often described as ``prognathic'' or underbite. 

\setcounter{figure}{0}
\renewcommand{\thefigure}{S\arabic{figure}}

\begin{figure}[!h]
    \centering

    \begin{subfigure}[t]{0.2\textwidth}
        \centering
        \includegraphics[width=\linewidth]{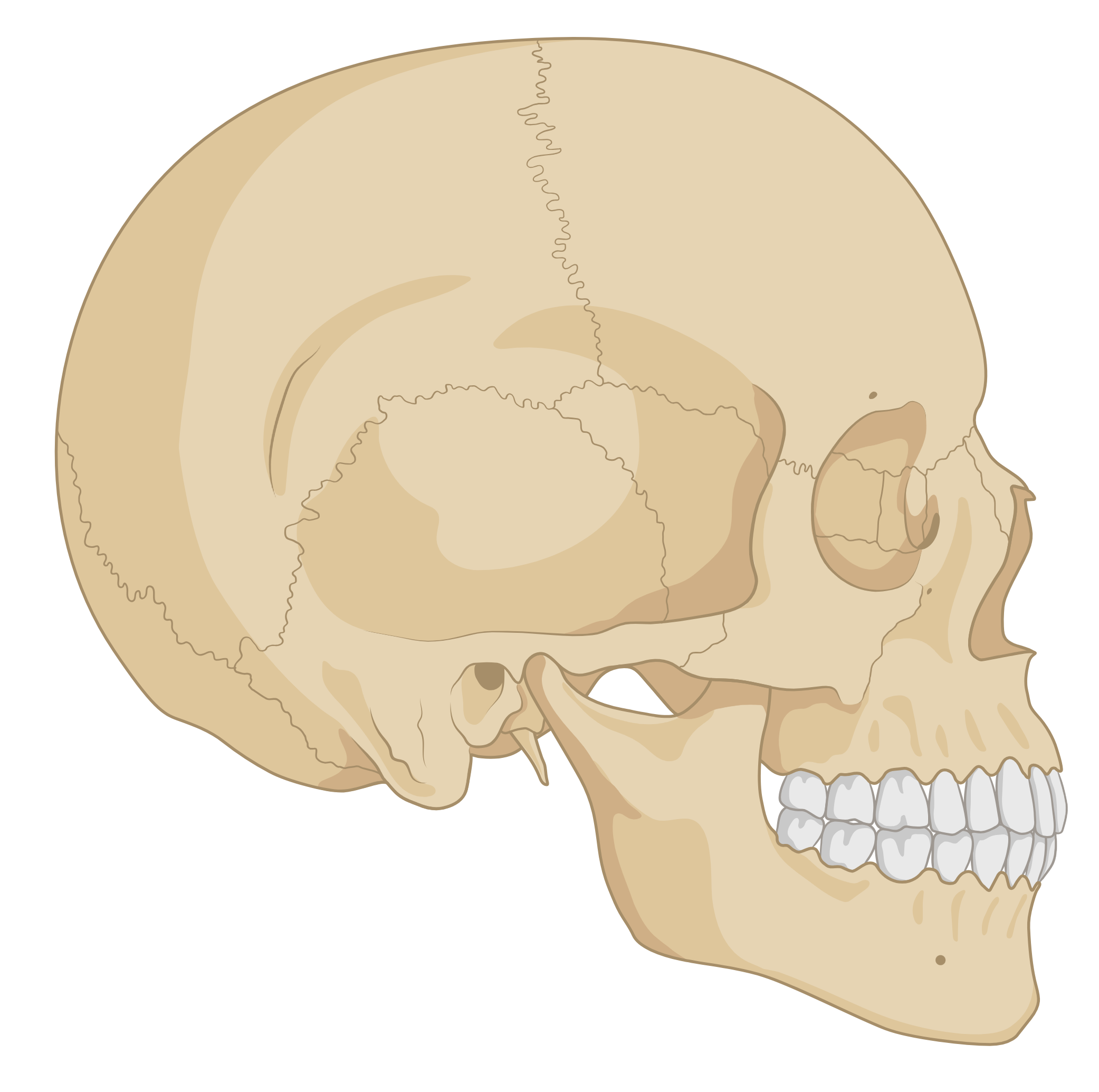}
        \caption{Class I}
    \end{subfigure}
    \hspace{2mm}
    \begin{subfigure}[t]{0.2\textwidth}
        \centering
        \includegraphics[width=\linewidth]{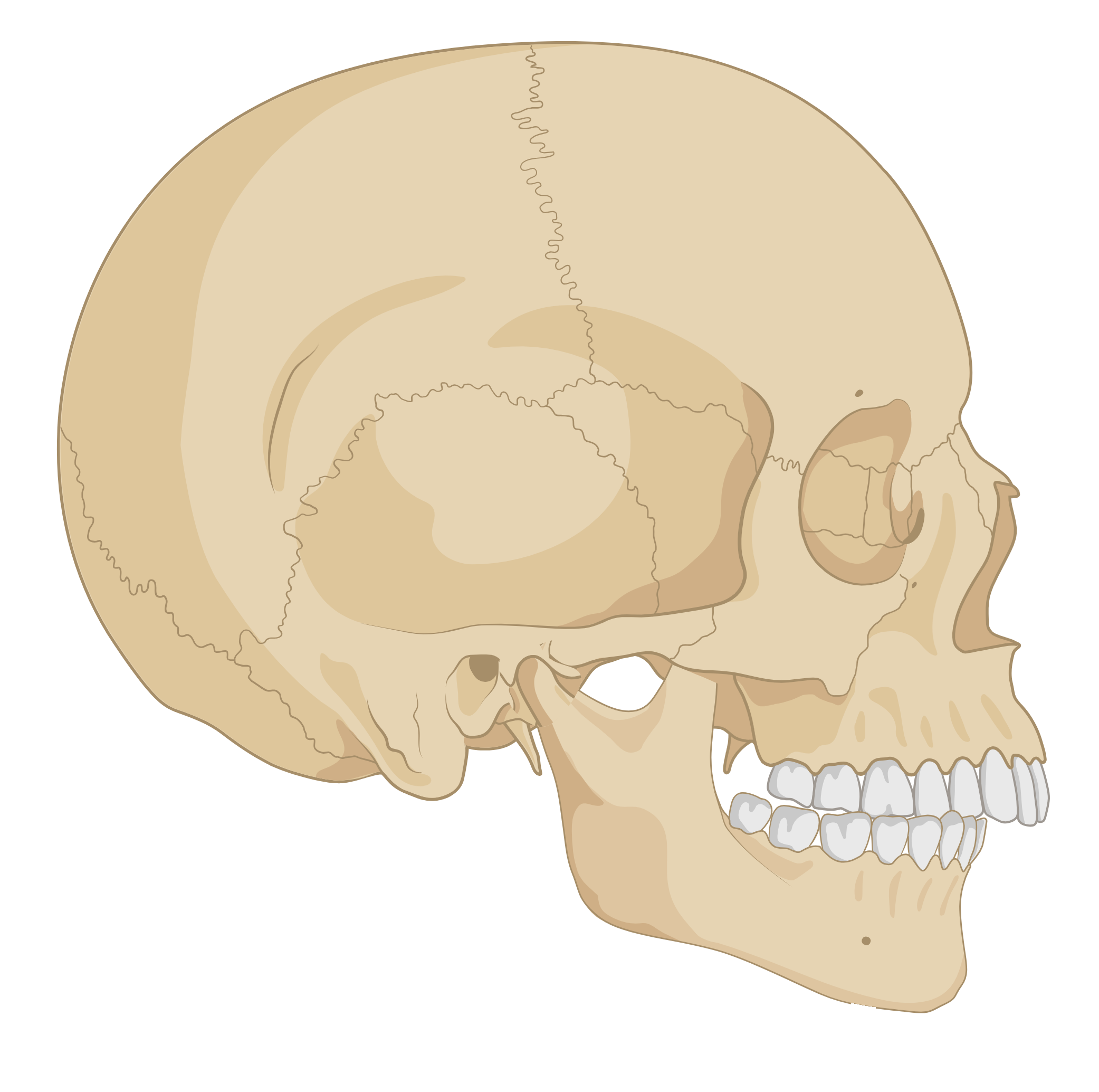}
        \caption{Class II}
    \end{subfigure}
    \hspace{2mm}
    \begin{subfigure}[t]{0.2\textwidth}
        \centering
        \includegraphics[width=\linewidth]{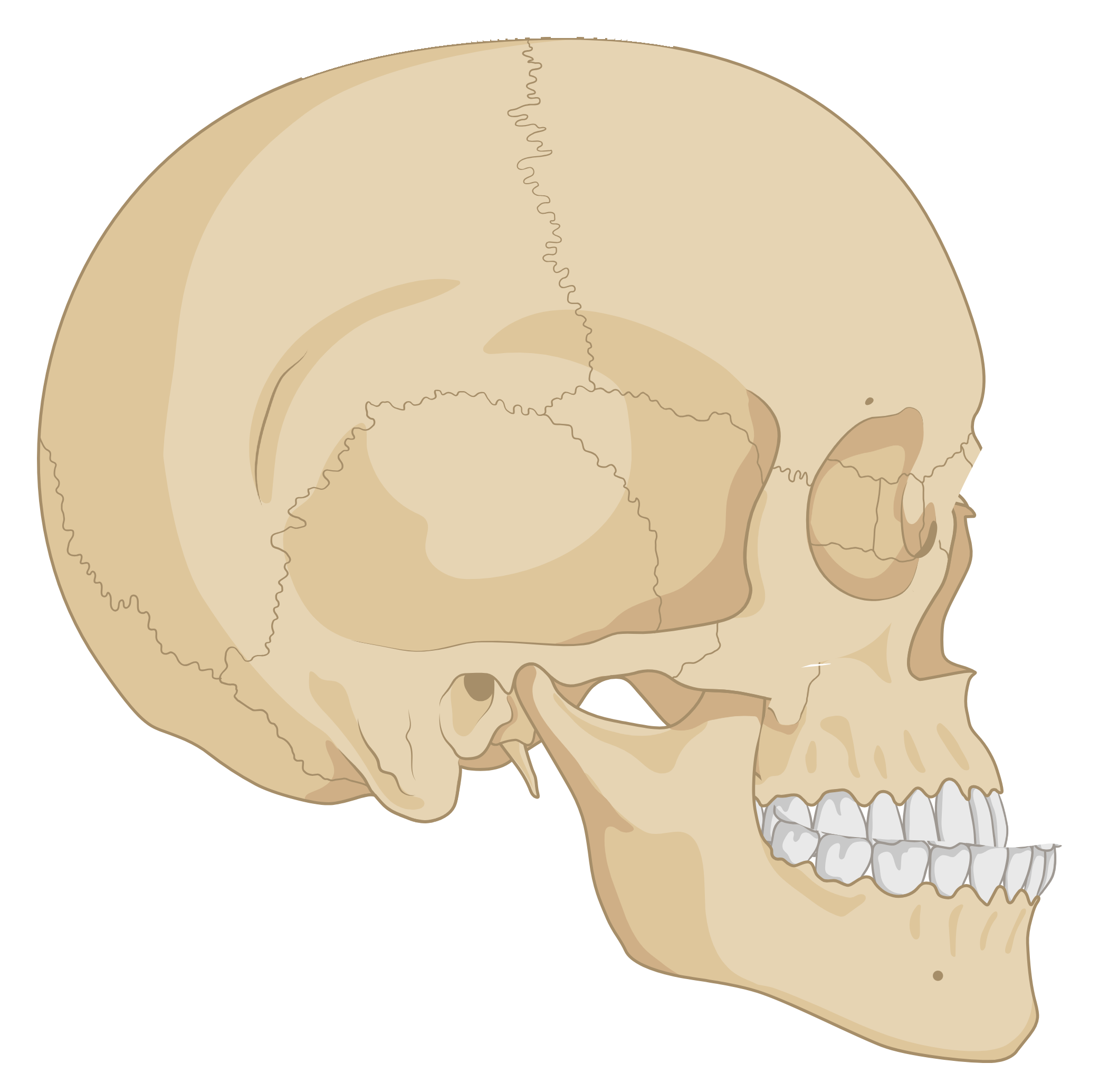}
        \caption{Class IIIA}
    \end{subfigure}
    \hspace{2mm}
    \begin{subfigure}[t]{0.2\textwidth}
        \centering
        \includegraphics[width=\linewidth]{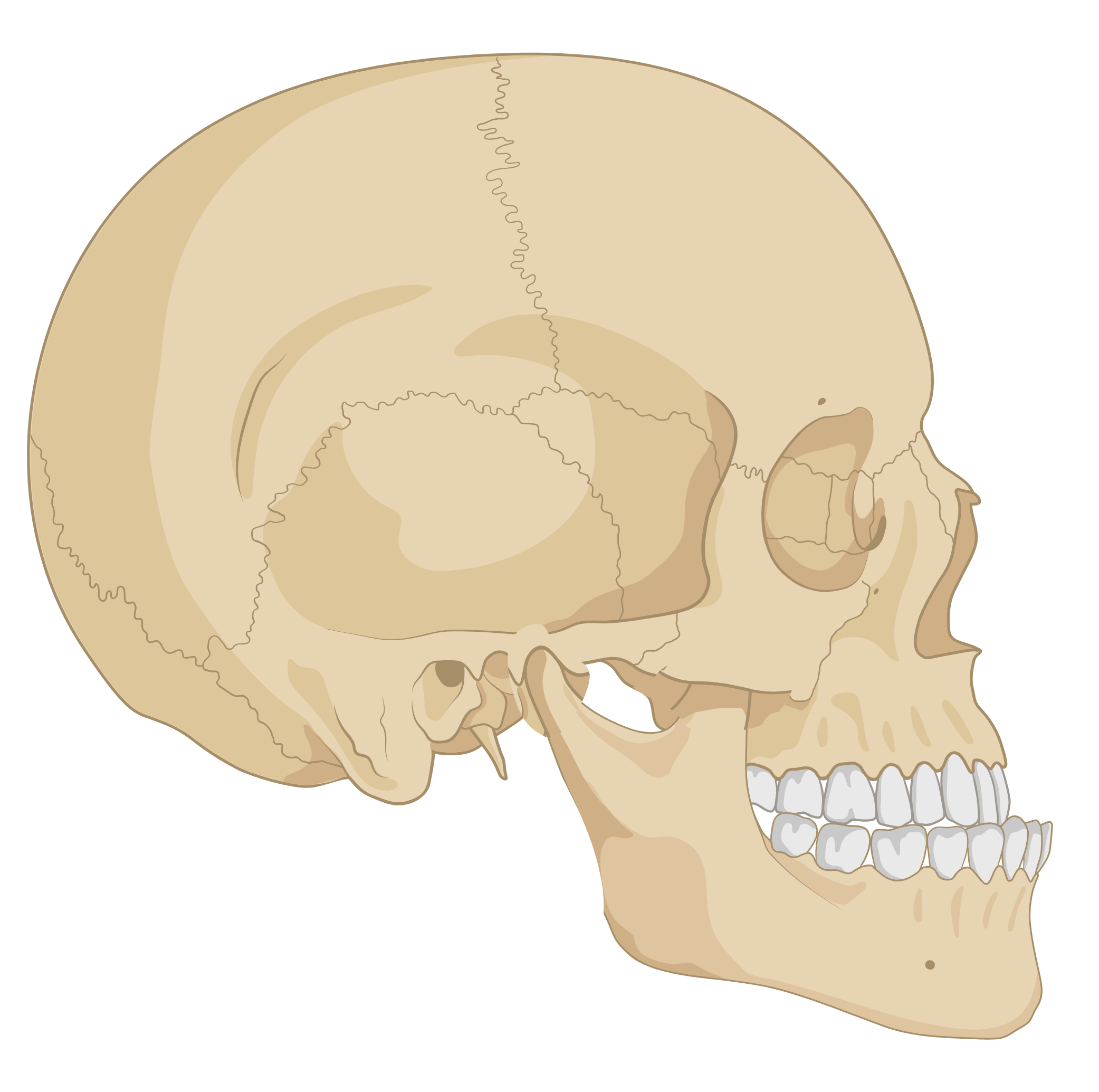}
        \caption{Class IIIB}
    \end{subfigure}

    \caption{Craniofacial malocclusion classes}
    \label{fig:skeletal-classification}
\end{figure}

The Class III classification was expanded into two subgroups to more accurately describe the skeletal discrepancy with the use of two separate classifications, depending on the etiology of malocclusion: a Class IIIA classification to describe the cases where the more posterior position of the maxilla is the main contributing factor to the malocclusion, and a Class IIIB classification to describe the case where a more anterior position of the mandible is the main component. This subclassification is clinically significant and contributes to treatment planning decisions. In the case of Class II, the differential diagnosis related to the contribution of either jaw in the skeletal discrepancy is not as clinically critical, and, therefore, no Class II subgroups were created. 

\par\vspace{1em}

\begingroup
\setbox0=\hbox{[A]\ }
\noindent\hangindent=\wd0\hangafter=1
\makebox[\wd0][l]{[A]\ }Tang, E. L. K., \& Wei, S. H. Y. (1993). Recording and measuring malocclusion: a review of the literature. \textit{American Journal of Orthodontics and Dentofacial Orthopedics}, \textit{103}(4), 344--351.
\par
\endgroup

\end{document}